\documentclass[conference,compsoc]{IEEEtran}
\usepackage{amssymb,amsmath,epsfig,latexsym,graphicx,bm,multirow}
\usepackage{epstopdf}

\ifCLASSOPTIONcompsoc
  \usepackage[nocompress]{cite}
\else
  \usepackage{cite}
\fi
\ifCLASSINFOpdf
\else
\fi

\begin{document}
%
\title{CNN-Based Prediction of Frame-Level Shot Importance for Video Summarization}



%

\author{\IEEEauthorblockN{Mohaiminul Al Nahian\IEEEauthorrefmark{1},
A. S. M. Iftekhar\IEEEauthorrefmark{1}, Mohammad Tariqul
Islam\IEEEauthorrefmark{1} \\ S. M. Mahbubur
Rahman\IEEEauthorrefmark{2}, and Dimitrios Hatzinakos
\IEEEauthorrefmark{3}}
\IEEEauthorblockA{\IEEEauthorrefmark{1}Department of EEE,
Bangladesh University of Engineering and Technology, Dhaka 1205,
Bangladesh}\IEEEauthorblockA{\IEEEauthorrefmark{2}Department of
EEE, University of Liberal Arts Bangladesh, Dhaka 1209,
Bangladesh} \IEEEauthorblockA{\IEEEauthorrefmark{3} Department of
ECE,
University of Toronto, Toronto, ON, Canada, M5S 2E4\\
Email: nahian.3735@gmail.com, iftekharniloyeee@gmail.com,
tariqul@eee.buet.ac.bd\\mahbubur.rahman@ulab.edu.bd,
dimitris@comm.utoronto.ca}}


\maketitle

\begin{abstract}
In the Internet, ubiquitous presence of redundant, unedited, raw
videos has made video summarization an important problem.
Traditional methods of video summarization employ a heuristic set
of hand-crafted features, which in many cases fail to capture
subtle abstraction of a scene. This paper presents a deep learning
method that maps the context of a video to the importance of a
scene similar to that is perceived by humans. In particular, a
convolutional neural network (CNN)-based architecture is proposed
to mimic the frame-level shot importance for user-oriented video
summarization. The weights and biases of the CNN are trained
extensively through off-line processing, so that it can provide
the importance of a frame of an unseen video almost
instantaneously. Experiments on estimating the shot importance is
carried out using the publicly available database TVSum50. It is
shown that the performance of the proposed network is
substantially better than that of commonly referred feature-based
methods for estimating the shot importance in terms of mean
absolute error, absolute error variance, and relative $F$-measure.
\end{abstract}


%


\section{Introduction}
\begin{figure}[t]
\centering
\includegraphics[width=8cm,height=4cm]{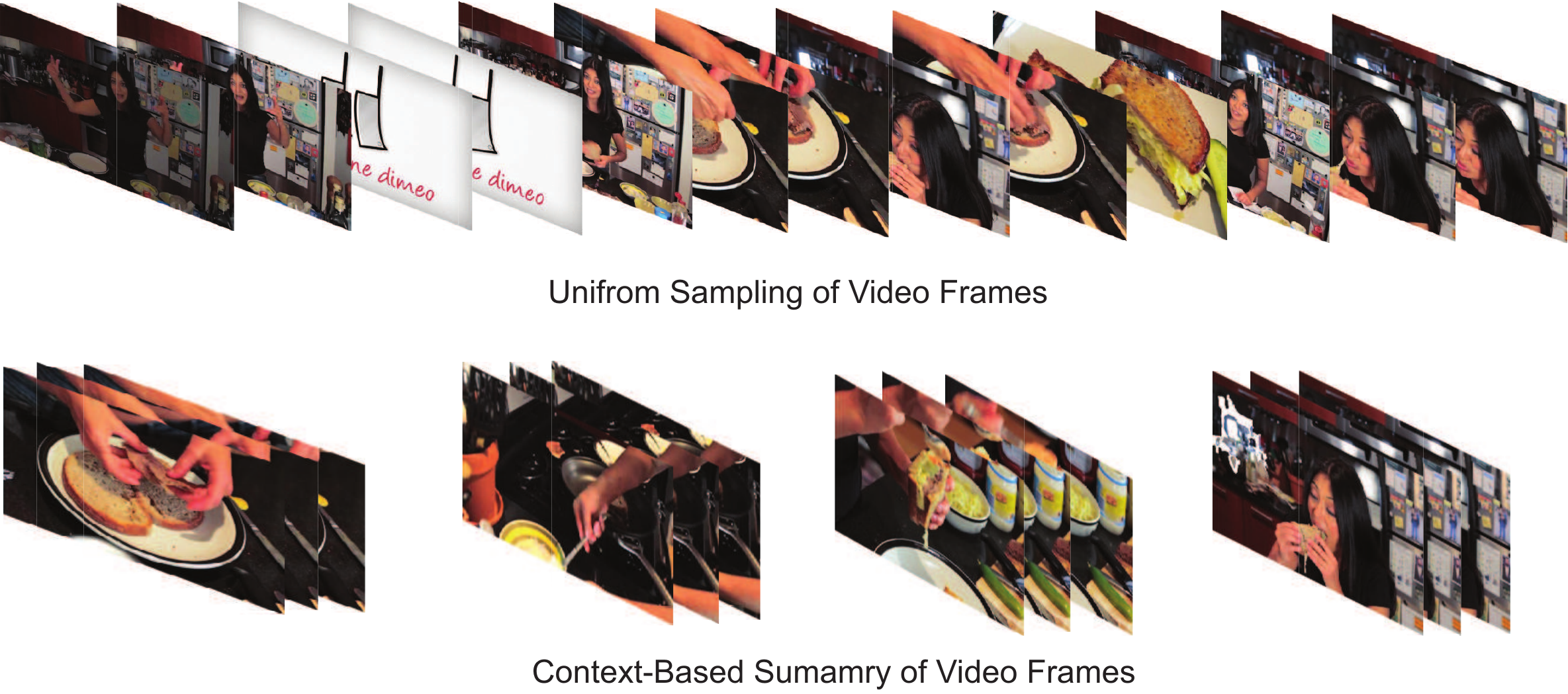} \caption{An illustrative
example showing effectiveness of context-based summarization
instead of uniform sampling of frames to review a
video.}\label{Fig:summar}
\end{figure}

With the development of comfortable and user-friendly devices for
capturing and storing multimedia content, a huge amount of videos
are being shot at every moment. Nearly 60 hours worth of footage
is uploaded on YouTube in every minute~\cite{brain_youtube_2014}.
To find and analyze this huge amount of videos have become an
extremely tedious task.
The generation of a compact, comprehensive, and automated summary
of video can facilitate an effective way to utilize videos for
various real-life applications such as for classifying huge number
of online videos, removing redundant videos, highlighting the
sports matches or trailer of feature films. Also, a semantical
relevant position can be located using video summaries that can be
essential for surveillance system~\cite{money_video_2008}.
Fig.~\ref{Fig:summar} shows an illustrative example of summary of
a video titled ``Reuben Sandwich with Corned Beef \& Sauerkraut''
available in YouTube.  A number of frames of the video are grouped
together based on the noticeable contexts as a summary of the
video. It is evident from Fig.~\ref{Fig:summar} that a
context-dependent video summary can have a better representation
as compared to uniform sampling of video frames.

\subsection{Related Works}
In general, there are three major approaches to summarize
videos~\cite{ding_study_1998}. They are:  object-based,
event-based and feature-based methods. Object-based approach
mainly depends on detecting the highlighted objects. The
underlying assumption is that these objects are the key elements
for a summary. In other words, the frames in which these objects
are found can be considered as the important frames to be
presented in the summary~\cite{meng_keyframes_2016}. Lee {\emph{et
al.}}~\cite{lee_discovering_2012} used these object-based
detection of key frames to summarize egocentric videos. Though
this approach is effective for certain types of videos, the
success of the methods largely depends on the content of the
videos. If a highlighted object is not present throughout the
entire video or the highlighted object is present in every frame
of a video, then object-based detection methods will not be able
to summarize the video effectively.

In the event-based methods, an important event is determined by
the use of previously defined bag-of-words. The events can be
detected by the change in various low-level factors, e.g., change
in colors or abrupt change in camera direction. These methods are
used by many works in the literature, when the goal as well as the
environments of summarization is very much specific, e.g.,
surveillance videos~\cite{lin_summarizing_2015}, sports
videos~\cite{li_event_2001}, coastal area
videos~\cite{cullen_detection_2012}. This approach fails to
represent the overall generality as similar events can have
contrasting significance in different environments. For example,
in a video of a football match, the scene of scoring goal is
considered important, but similar event in a surveillance video
can be useless.

Most popular methods for video summarization are based on suitable
features. In this approach, certain features are used to detect
important frames termed as key frames. In most cases, a large
number of features are combined together to detect important
frames. These features are selected by judging the content of the
videos. Different types of features including the visual
attention~\cite{ejaz_efficient_2013} and singular value
decomposition (SVD)~\cite{ntalianis_optimized_2005} have been used
for key frame detection. Recently, machine learning techniques
have been introduced to select suitable
features~\cite{lu_bag--importance_2014}. However, the success of
such methods seriously depends on the number of selected features,
and the way the features are combined. Hence, the methods fail to
map individual perception in a generalized framework.

\subsection{Scope of Work}
Most of the existing methods for video summarization focus on
detecting key frames based on some sorts of fixed parameters. This
type of parameter-based detection is not suitable for an overall
general platform of video summarization. In this work, a
convolutional neural network (CNN)-based architecture is proposed
to deal with the overall generality of the problem and to estimate
the importance of each frame in a video. This can be used to
develop a platform in which a user can have freedom to select the
length of the summary as applicable. To the best of our knowledge,
finding out the frame-level shot importance using the CNN is not
present in the current literature.

\subsection{Specific Contributions}
The main objective of the paper is to present a CNN model to
estimate the shot-by-shot importance in a video. The overall
contributions of the paper are:
\begin{itemize}
\item Developing a CNN based algorithm to estimate frame-level
shot importance.

\item Generating a platform for the summarization of any kind of
video using the estimated frame-level shot importance.
\end{itemize}

The rest of the paper is organized as follows. Section~2 provides
a description of the proposed architecture. The experimental setup
and the results obtained are described in Section~3. Finally,
Section~4 provides the conclusion.

\begin{figure*}[t]
\centering
\includegraphics[width=12cm,height=5cm]{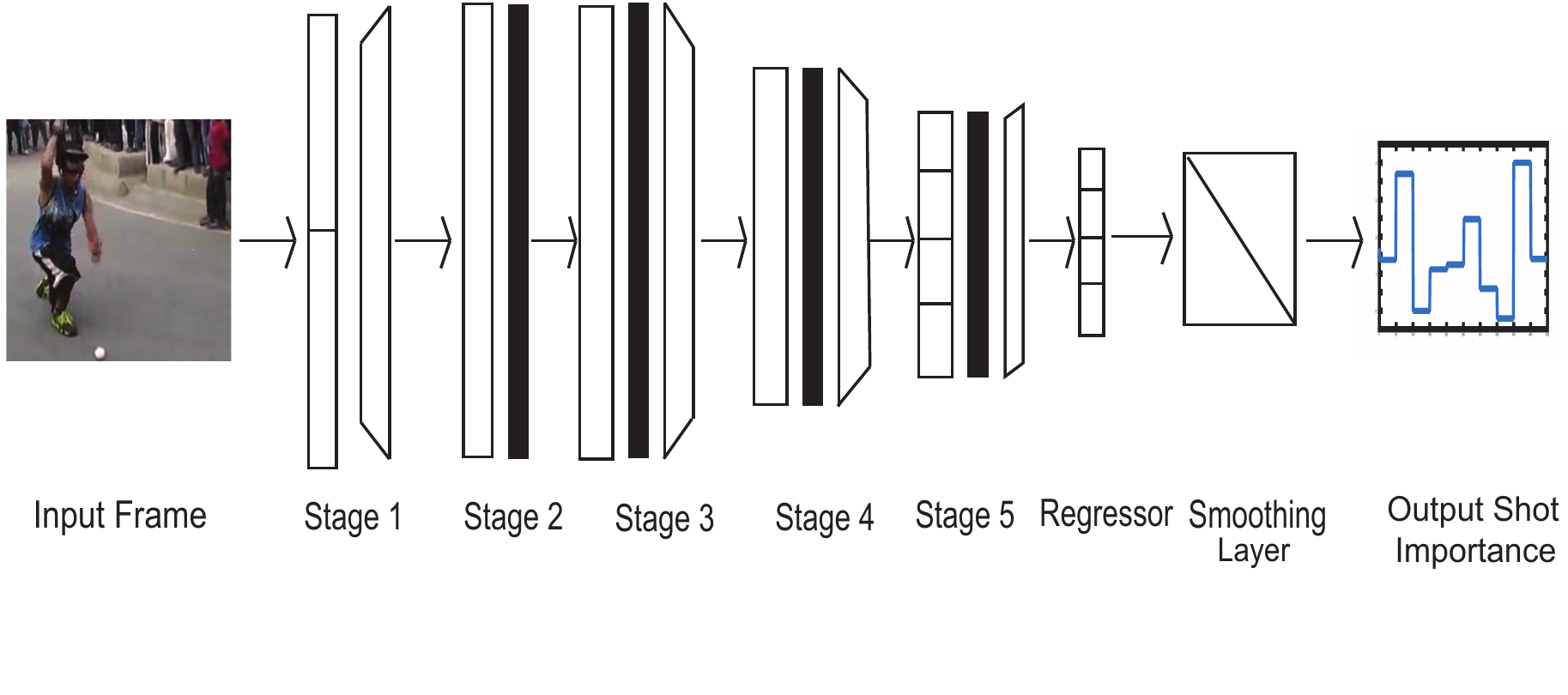} \caption{Proposed CNN
model for predicting the frame-level shot importance of a video.
The input to the model is raw video frames and output is the score
of importance.}\label{Fig:stick}
\end{figure*}

\section{Proposed Method}
In this paper, a feed-forward CNN is employed to determine the
frame-level shot importance of a video. In the proposed multilayer
CNN, the first layer is the input layer which uses the raw video
frame $\mathbf{X}_0$ as the input. The last layer is the output
layer that predicts the importance score $y$ $(0\leq y \leq L)$
$(y\in\mathbb{R})$ of the input frame in that particular video,
where $L$ is a positive integer corresponding to the highest
score. A low value of $y$ indicates a less important frame, while
a high value implies an important one. In this section, first the
proposed CNN model is described. Then, the training and
optimization schemes are detailed.

\subsection{CNN Model}\label{propmodel}
In order to estimate the shot importance of a frame, we train an
end-to-end CNN that automatically learns visual contexts to
predict the score in the output. The proposed CNN architecture is
a six-stage model employing learnable convolution and fully
connected layers as shown in the stick diagram in
Fig.~\ref{Fig:stick}. The convolution and fully connected
operations are followed by ReLU activation
for its ability to help neural networks attaining a better sparse
representation~\cite{glorot_deep_2011}. The first stage of the
network is pre-processing tasks needed to normalize the dimension
of the data. The pre-processing stage can be written as
\begin{align}
\mathbf{X}_1=\textrm{preprocess}(\mathbf{X_0})
\end{align}
This task involves frame resizing and cropping that are applied
sequentially. In the stick diagram of Fig.~\ref{Fig:stick} the
frame resizing is shown using a rectangle with a single stripe and
cropping operation is shown by a diverging trapezoid. The second
stage performs a convolution operation which employs ReLU
activation on $\mathbf{X}_1$, which is given by
\begin{align}
\mathbf{X}_2 =\max(\mathbf{0},\mathbf{W}_1\ast
\mathbf{X}_1+\mathbf{b}_1)
\end{align}
where $\ast$ is the convolution operation,
$\max(\mathbf{0},\cdot)$ is the ReLU operation. In
Fig.~\ref{Fig:stick}, the convolution layer is shown as a
rectangle and ReLU layer as a solid line. The third and fourth
stages use the convolution, ReLU and max-pooling operations
serially. These operations are given by the equations
\begin{align}
\mathbf{X}_3 =MP(\max(\mathbf{0}, \mathbf{W}_2 \ast \mathbf{X}_2+\mathbf{b}_2))\\
\mathbf{X}_4 =MP(\max(\mathbf{0}, \mathbf{W}_3  \ast
\mathbf{X}_3+\mathbf{b}_3))
\end{align}
where $MP(\cdot)$ is the max-pool operation. This operation
reduces the spatial dimension by half and is represented by
converging trapezoid (see Fig.~\ref{Fig:stick}). The fifth stage
consists of fully connected operation, the ReLU and dropout
layers. First, the output of fourth stage $\mathbf{X}_4$ is
flattened to a 1-D vector $\widehat{\mathbf{X}}_4$, and then this
vector is fed into the fifth stage to provide the output
$\mathbf{X}_r$ given by
\begin{equation}
\mathbf{X}_r =\textrm{Drop}(\max(\mathbf{0},
\mathbf{W}_4^T\widehat{\mathbf{X}}_4+\mathbf{b}_4))
\end{equation}
where $\textrm{Drop}(\cdot)$ is the dropout
operation~\cite{srivastava_dropout:_2014}. In
Fig.~\ref{Fig:stick}, the fully connected layer and the dropout
layer are represented by rectangle with three stripes in the
middle and a parallelogram, respectively. The final part of the
CNN is the regressor which is a fully connected layer that outputs
the estimation of the frame importance in scalars from
$\mathbf{X}_r$ given by
\begin{equation}
\hat{y} = \mathbf{W}_r^T \mathbf{X}_r+b_r
\end{equation}
In many cases, the frame-level importance can be averaged over a
few neighboring frames using a smoothing filter (shown as
rectangle with a diagonal stripe in Fig.~\ref{Fig:stick}).
Overall, the learnable parameters of the network are the filter
sets, $\mathbf{W}_1$, $\mathbf{W}_2$, $\mathbf{W}_3$,
$\mathbf{W}_4$, and $\mathbf{W}_r$, and their corresponding bias
terms $\mathbf{b}_1$, $\mathbf{b}_2$, $\mathbf{b}_3$,
$\mathbf{b}_4$ and $b_r$, respectively.

\subsection{Training and Optimization}
There are number of training and optimization schemes that can be
chosen for attaining good results from the network. An effective
choice of initialization of weights and biases can significantly
reduce training time by converging the network faster. In this
context, we have explored the works of Glorot \emph{et
al.}~\cite{glorot_understanding_2010} and initialized all the
biases with zeros and weights $\mathbf{W}_i$ at each layer by
taking samples from a uniform distribution
$\mathbf{W}_i\sim
\mathcal{U}\left[\frac{-1}{\sqrt{M}},\frac{1}{\sqrt{M}}\right]$
where $M$ $(M\in\mathbb{Z})$ is the size of the previous layer. In
order to apply back-propagation~\cite{rumelhart_learning_1988} for
training the network, a loss function is required to be specified
that is easily differentiable. For regression-based tasks such as
the estimation of scores, most common choices are $\ell_1$-norm,
$\ell_2$-norm or \emph{Frobenius} norm. In the proposed method, we
choose an $\ell_2$-norm-based loss function given by
\begin{align}
C=\sum_{n=1}^{N}||y_n-\hat{y}_n||^2
\end{align}
where $y_n$ is the ground truth value of the shot importance,
$\widehat{y}_n$ is the predicted score, and $N$ $(N\in\mathbb{Z})$
is the number of training inputs fed into the back-propagation
process in each iteration for mini-batch
optimization~\cite{li2014efficient}. During the training period,
this function is optimized by using the contemporary Adam
stochastic optimization technique~\cite{kingma_adam:_2014}. The
weights of the filter sets denoted by $w$ are updated based on the
first moment $\widehat{m}$ and second moment $\widehat{v}$ of the
gradient of the loss function $C$ with respect to the weights.
Overall, the update process of the optimization can be written
as~\cite{kingma_adam:_2014}
\begin{align}
\widehat{m}(t)&=\widehat{m}(t-1) + (1-\beta_1) \frac{dC(t)}{dw(t)}\\
\widehat{v}(t)&= \widehat{v}(t-1) + (1-\beta_2)\left(\frac{\partial C(t)}{\partial w(t)} \right)^2\\
w(t)&=w(t-1)-\frac{{\alpha{\widehat{m}(t)}}}{{\sqrt{\widehat{v}(t)}+\epsilon}}
\end{align}
where $\alpha$ $(\alpha>0)$ is the step size, $\beta_1$ and
$\beta_2$ $(\beta_1,\beta_2>0)$ are decay rates for the first and
second moments, respectively, and $\epsilon$ $(\epsilon>0)$ is a
factor of numerical stability.

\section{Experiments and Results}
Experiments are carried out to evaluate the performance of the
proposed CNN architecture as compared to existing methods for
predicting the score of frame importance in videos. In this
section, first we give an overview of video dataset used in the
experiments, then we describe our training and testing data
partitions, data augmentation techniques, parameter settings of
the proposed architecture, and matching scheme of estimated score
of importance with the ground truth. Then, the methods compared
for performance evaluation are introduced. Finally, results are
presented and evaluated in terms of commonly-referred performance
metrics of regression.

\subsection{Database}
In the experiments, we have used the TVSum50
database~\cite{song_tvsum:_2015} that includes 50 video sequences.
These videos are categorized into ten different genres including
the flash mob, news, and video blog. Each genre contains videos of
five independent scene. The duration of videos varies from 2 to 10
minutes. Each frame of these videos has been annotated by an
importance score of continuous values ranging from 1 to 5 by using
crowd-sourcing. It is found empirically that a shot length of two
seconds will be able to the reflect local context of a
video~\cite{song_tvsum:_2015}. By adopting this rule, each video
is divided into segments, where each segment has a duration of two
seconds. These segments are first annotated by 20 users. A ground
truth of importance score has been produced by regularizing and
combining these annotated scores.

\begin{figure*}[t]
\centering
\includegraphics[width=18cm,height=10cm]{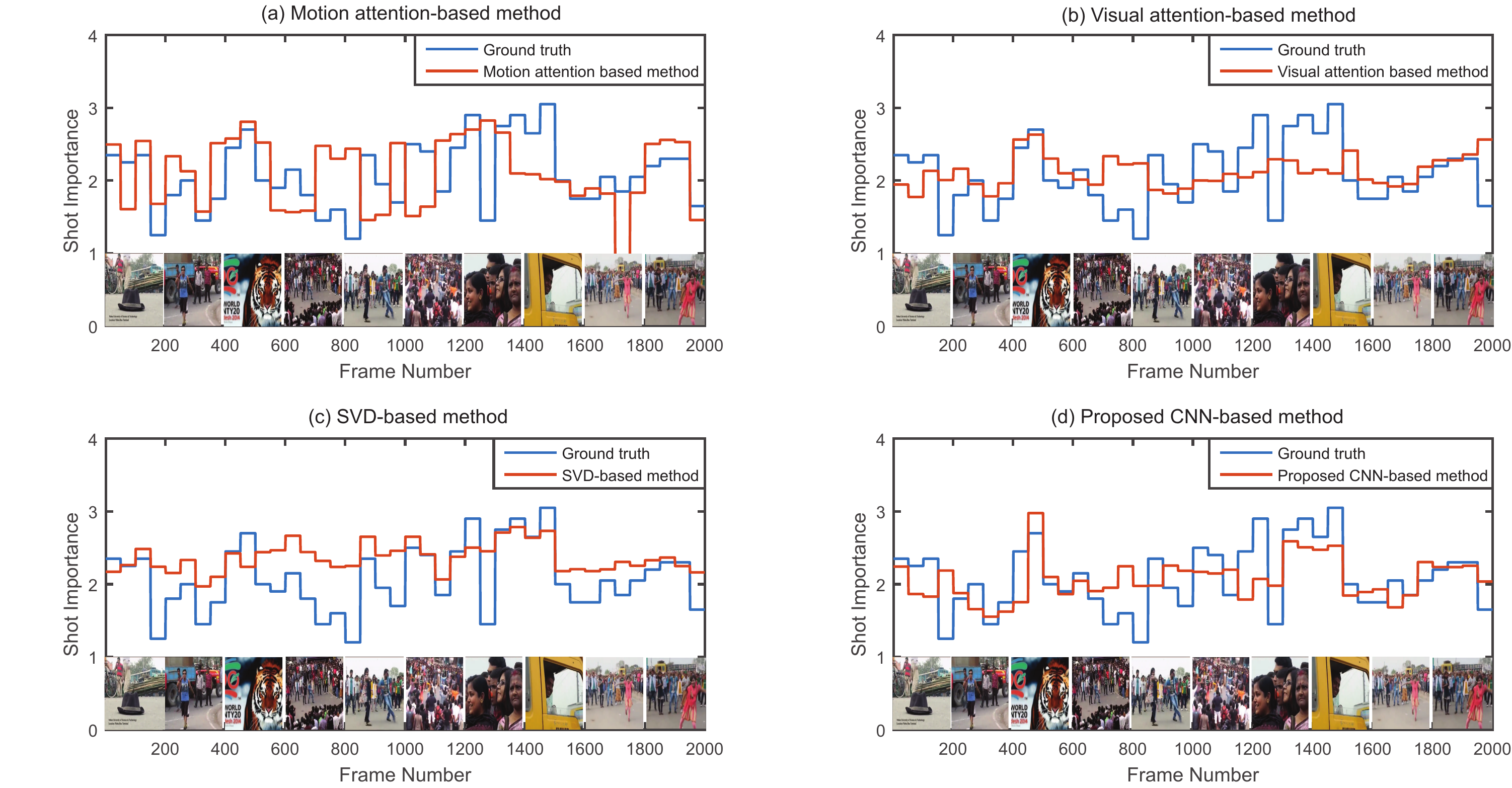} \caption{Frame-level
scores of shot importance predicted by using the experimental
methods. The predicted scores are compared with the ground truth.
The comparisons are shown for (a) motion attention-based method,
(b) visual attention-based method, (c) SVD-based method, and (d)
proposed CNN-based method.}\label{output}
\end{figure*}

\subsection{Setup}\label{sec:setup}
Out of 50 videos of the dataset, 35 videos are chosen for training
and the mutually exclusive rest of the 15 videos are kept for
testing phase. In order to design a fair evaluation process, at
least three videos for the training set and one video for testing
set are included from each of the ten genres. In order to achieve
a computational efficiency and to reduce the training period, a
subset of frames from videos are considered for learning. In
particular, a single frame from each strip of five consecutive
frames is considered for training scheme. This is mainly due to
the fact that the visual contents of five consecutive frames are
almost same in a video. This ensures that the training data has
less amount of redundant information and, thus the approach
significantly reduces the training period. On the other hand, no
frames is discarded from the test set, instead the importance
score of every frame of a video is predicted.

\subsection{Data Augmentation}\label{augment}
Data augmentation helps to achieve generalized results in
CNN-based learning~\cite{krizhevsky_imagenet_2012}. It reduces
overfitting by virtually increasing the training data size. In
general, a larger network can be trained by augmenting a dataset
without losing validation accuracy. This scheme has been adopted
in our experiments. Augmentation techniques that are used in the
training include the transpose, horizontal flips, and vertical
flips of the frames. One or more of these operations are chosen
randomly in each stage of the training step. In other words, seven
new variants of the original data are achieved, and our training
set virtually increases by up to 8 times. During each iteration, a
random integer is generated between 1 and 8 inclusive that
correspond to a specific combination of data augmentation
techniques. Based on the generated integer, the selected
operations are performed on the data prior to feeding it to the
following stage.

\subsection{Parameter Settings}
The network parameters of the CNN model described in
Section~\ref{propmodel} are chosen based on the dimensions of
input and required output in different layers. Since the size of
input video frames varies among different videos, first the video
frames are resized to $284 \times 284 \times 3$ and then cropped
centrally to obtain $256 \times 256 \times 3$ sized images, where
$3$ is the channel parameter of RGB components of a color image.
The number of filters in the sets $\mathbf{W_1}$, $\mathbf{W_2}$,
$\mathbf{W_3}$, $\mathbf{W_4}$, and $\mathbf{W_r}$ and
corresponding number of bias terms $\mathbf{b_1}$, $\mathbf{b_2}$,
$\mathbf{b_3}$, $\mathbf{b_4}$ and $b_r$ are set to $32$, $64$,
$64$, $10$ and $1$, respectively, since such a choice provides an
overall good performance. The kernel size of all the convolution
filters is set to $5$ and that of the max-pool operation is set to
$2$. The dropout parameter is chosen as $0.5$ during training and
$1$ during testing. Empirically the parameters  $\alpha$,
$\beta_1$, $\beta_2$ of the Adam optimizer are found to be
$10^{-4}$, $0.9$ and $0.999$, respectively. The numerical
stability factor $\epsilon$ is set to $10^{-8}$.

\subsection{Matching of Importance}\label{smooth}
A single value has been assigned as the shot importance for 50
neighboring frames in the ground truth. Since the proposed model
predicts shot importance for each of the frames in a video, a
scheme for matching the importance has been employed in order to
be consistent with the ground truth of dataset. In particular,
first the predicted output values for 50 consecutive frames are
considered, then the minimum $10\%$ and maximum $10\%$ of the
predictions are discarded, and finally the root mean squared (RMS)
value of the remaining data is assigned as the fixed-level shot
importance for the 50 neighboring frames.


\begin{table*}[!ht]
    \renewcommand{\arraystretch}{1.5}   \centering
    \caption{Performance of Prediction of Shot Importance in
Terms of MAE, AEV and Relative $F$-measure}
    \begin{tabular}{|p{3 cm}|p{2.5 cm}|p{2.5 cm}|p{2.5 cm}|}
        \hline              \centering
        Methods & MAE & AEV & Relative $F$-measure \\
        \hline
        Motion Attention~\cite{ejaz_efficient_2013} & $0.4791$ & $0.1280$ & $0.6018$ \\
        \hline
        Visual Attention~\cite{ma_user_2002} & $0.3842$ & $0.2282$ & $0.3679$\\
        \hline
        SVD~\cite{ntalianis_optimized_2005} & $0.3639$ & $0.0808$ & $0.6915$\\
        \hline
        Proposed CNN & $\mathbf{0.3212}$ & $\mathbf{0.0572}$ & $\mathbf{0.7222}$\\
        \hline  \end{tabular}   \label{tebil}
\end{table*}

\subsection{Comparing Methods}
The proposed CNN is a learning-based method, where the importance
of frames are predicted automatically by the network. In the
experiments, we select three feature-based approaches reported for
video summarization. Originally the methods are concerned with the
selection of key frames. The methods are briefly described as
follows:

\begin{itemize}

\item Visual attention~\cite{ejaz_efficient_2013}: In this method,
the visual attention extracted from spatial and temporal saliency
is used to extract key frames from a video.

\item Motion attention~\cite{ma_user_2002}: The video features
extracted from motions are employed for video summarization.

\item Singular value decomposition
(SVD)~\cite{ntalianis_optimized_2005}: The minimization of cross
correlation of the features extracted in terms of SVD of frames is
used to identify the key frames for video summarization.

\end{itemize}

To compare these methods with the proposed one, they are invoked
to predict shot importance for each of the frames of a video. In
particular, the features are used in a support vector regression
technique to predict the frame-level shot importance using the
same training and testing sets described in
Section~\ref{sec:setup}.

\subsection{Evaluation Metrics}
The performance of the proposed CNN-based method and three
comparing methods are evaluated in terms of three metrics, namely,
mean absolute error (MAE), absolute error variance (AEV), and
relative $F$-measure. The MAE indicates how much the predicted
values deviate from the ground truth on average, and the AEV
reveals the fluctuations of absolute errors. Thus, a lower value
of MAE means predicted value is very close to the actual one.
Similarly, a small AEV is a good sign implying that errors do not
fluctuate significantly.

The $F$-measure gives an idea about the close matching between the
video summary prepared by the predicted shot importance and that
by the ground truth. In order to compute the $F$-measure, a
threshold is selected for each of the comparing methods as well as
for the ground truth. The threshold maps the continuous values of
frame importance into binary values denoting the selected and
non-selected frames for a summary, preferably with a length of
$5\%-15\%$ of the original video. The metric $F$-measure is given
by
\begin{equation}
F\textrm{-measure}=2 \times \frac{{Precision \times
{Recall}}}{{Precision+Recall}}
\end{equation}
where $Precision$ is the fraction of matched frames with respect
to the ground truth, and $Recall$ implies the fraction of matched
frames with respect to the total number of frames. To find out how
well the proposed CNN-based method performs as compared to others,
the relative $F$-measure is evaluated by normalizing the metric
with the same calculated from the annotated ground truths of
fifteen videos.

\subsection{Results}
In the experiments, shot importance of all the frames of test
videos are predicted using the proposed as well as three comparing
methods. Then, the importance values are grouped for local
neighboring 50 frames as described in Section~\ref{smooth}.
Table~\ref{tebil} shows the overall prediction performance of the
testing videos in terms of the metrics MAE, AEV and relative
$F$-measure. It is seen from the table that the proposed CNN-based
method performs the best by providing the lowest MAE. It shows
approximately $13\%$ improvement in terms of MAE from the most
competitive method reported in~\cite{ntalianis_optimized_2005},
which uses SVD of frames as features. The proposed method
outperforms the comparing methods by showing an improvement of at
least $40\%$ in robustness by providing the lowest AEV. It can
also be found from Table~\ref{tebil} that our method provides the
highest relative $F$-measure as compared to others, where the
improvement is more than $4\%$ from the competing method. In other
words, our proposed method performs significantly better than
others for predicting the shot importance. This is evident because
the method consistently provides low absolute errors through out
the entire frames of a video and thus results in a video
summarization close to the ground truth.

Fig.~\ref{output} shows the frame-level scores of shot importance
predicted for first two thousand frames of a test video with flash
mob genre having a title of \emph{``ICC World Twenty20 Bangladesh
2014 Flash Mob Pabna University of Science \& Technology (PUST)"}.
This video was shot by a group of Bangladeshi students as a
promotional video of the 2014 ICC World Twenty20 event. It is seen
from Fig.~\ref{output} that the predicted scores of importance
provided by the proposed CNN-based method tend to follow the
ground truth more closely than that provided by the comparing
three methods. The motion-based method~\cite{ma_user_2002} shows
sudden changes of scores of importance, which appear even in the
opposite direction to the trend of the ground truth. The visual
attention~\cite{ejaz_efficient_2013} and
SVD-based~\cite{ntalianis_optimized_2005} methods though follow
the trend of ground truth closely in a few region, the deviations
are significant in most of the regions. Evidently, the above two
limitations are nearly absent in the prediction scores of the
proposed method, and hence, the CNN-based prediction appears to be
accurate and robust.

\section{Conclusion}
In this paper, a CNN-based architecture has been proposed to
predict frame-level shot importance of videos. The predicted
scores of shot importance can be used for the development of a
platform, which can provide a user-oriented automated summary of a
video. Thus, our work successfully converts the subjective video
summarization into a measurable objective framework. To evaluate
the proposed CNN-based method, annotated importance of ten genres
of videos of TVSum50 database have been used as the ground truth.
Experiments have been conducted by adopting mutually exclusive
training and testing sets that encompasses available genres of the
dataset. The proposed method has been compared with the methods
based on the visual attention, motion attention, and SVD features.
Experimental results reveal that the proposed CNN-based method
outperforms the existing feature-based methods in terms of three
evaluation metrics, namely, MAE, AEV and relative $F$-measure.

\bibliographystyle{IEEEtran}

\begin{thebibliography}{}
\providecommand{\url}[1]{#1}
\csname url@samestyle\endcsname
\providecommand{\newblock}{\relax}
\providecommand{\bibinfo}[2]{#2}
\providecommand{\BIBentrySTDinterwordspacing}{\spaceskip=0pt\relax}
\providecommand{\BIBentryALTinterwordstretchfactor}{4}
\providecommand{\BIBentryALTinterwordspacing}{\spaceskip=\fontdimen2\font plus
\BIBentryALTinterwordstretchfactor\fontdimen3\font minus
  \fontdimen4\font\relax}
\providecommand{\BIBforeignlanguage}[2]{{%
\expandafter\ifx\csname l@#1\endcsname\relax
\typeout{** WARNING: IEEEtran.bst: No hyphenation pattern has been}%
\typeout{** loaded for the language `#1'. Using the pattern for}%
\typeout{** the default language instead.}%
\else
\language=\csname l@#1\endcsname
\fi
#2}}
\providecommand{\BIBdecl}{\relax}
\BIBdecl

\end{thebibliography}


\begin{thebibliography}{10}
\providecommand{\url}[1]{#1} \csname url@rmstyle\endcsname
\providecommand{\newblock}{\relax}
\providecommand{\bibinfo}[2]{#2}
\providecommand\BIBentrySTDinterwordspacing{\spaceskip=0pt\relax}
\providecommand\BIBentryALTinterwordstretchfactor{4}
\providecommand\BIBentryALTinterwordspacing{\spaceskip=\fontdimen2\font
plus \BIBentryALTinterwordstretchfactor\fontdimen3\font minus
  \fontdimen4\font\relax}
\providecommand\BIBforeignlanguage[2]{{%
\expandafter\ifx\csname l@#1\endcsname\relax
\typeout{** WARNING: IEEEtran.bst: No hyphenation pattern has been}%
\typeout{** loaded for the language `#1'. Using the pattern for}%
\typeout{** the default language instead.}%
\else \language=\csname l@#1\endcsname \fi #2}}

\bibitem{brain_youtube_2014}
S.~Brain, ``{YouTube} statistics,'' \emph{Retrieved from statistic
brain:
  http://www.statisticbrain.com/youtube-statistics}, 2014.

\bibitem{money_video_2008}
A.~G. Money and H.~Agius, ``Video summarisation: {A} conceptual
framework and
  survey of the state of the art,'' \emph{J. Visual Communication and Image
  Representation}, vol.~19, no.~2, pp. 121--143, 2008.

\bibitem{ding_study_1998}
W.~Ding and G.~Marchionini, ``A study on video browsing
strategies,''
  University of Maryland at College Park, College Park, MD, Tech. Rep. Report
  No. UMIACS-TR-97-40, 1998.

\bibitem{meng_keyframes_2016}
J.~Meng, H.~Wang, J.~Yuan, and Y.-P. Tan, ``From keyframes to key
objects:
  {Video} summarization by representative object proposal selection,'' in
  \emph{Proc. IEEE Conf. Computer Vision and Pattern Recognition}, Las Vegas,
  NV, 2016, pp. 1039--1048.

\bibitem{lee_discovering_2012}
Y.~J. Lee, J.~Ghosh, and K.~Grauman, ``Discovering important
people and objects
  for egocentric video summarization,'' in \emph{Proc. IEEE Conf. Computer
  Vision and Pattern Recognition}, Providence, RI, 2012, pp. 1346--1353.

\bibitem{lin_summarizing_2015}
W.~Lin, Y.~Zhang, J.~Lu, B.~Zhou, J.~Wang, and Y.~Zhou,
``Summarizing
  surveillance videos with local-patch learning-based abnormality detection,
  blob sequence optimization, and type-based synopsis,'' \emph{Neurocomputing},
  vol. 155, pp. 84--98, 2015.

\bibitem{li_event_2001}
B.~Li and M.~I. Sezan, ``Event detection and summarization in
sports video,''
  in \emph{Proc. IEEE Work. Content-Based Access of Image and Video Libraries},
  Kauai, HI, 2001, pp. 132--138.

\bibitem{cullen_detection_2012}
D.~Cullen, J.~Konrad, and T.~D. Little, ``Detection and
summarization of
  salient events in coastal environments,'' in \emph{IEEE Int. Conf. Advanced
  Video and Signal-Based Surveillance}, Beijing, China, 2012, pp. 7--12.

\bibitem{ejaz_efficient_2013}
N.~Ejaz, I.~Mehmood, and S.~W. Baik, ``Efficient visual attention
based
  framework for extracting key frames from videos,'' \emph{Signal Processing:
  Image Communication}, vol.~28, no.~1, pp. 34--44, 2013.

\bibitem{ntalianis_optimized_2005}
K.~S. Ntalianis and S.~D. Kollias, ``An optimized key-frames
extraction scheme
  based on {SVD} and correlation minimization,'' in \emph{IEEE Int. Conf.
  Multimedia and Expo.}, Amsterdam, The Netherlands, 2005, pp. 792--795.

\bibitem{lu_bag--importance_2014}
S.~Lu, Z.~Wang, T.~Mei, G.~Guan, and D.~D. Feng, ``A
bag-of-importance model
  with locality-constrained coding based feature learning for video
  summarization,'' \emph{IEEE Trans. Multimedia}, vol.~16, no.~6, pp.
  1497--1509, 2014.

\bibitem{glorot_deep_2011}
X.~Glorot, A.~Bordes, and Y.~Bengio, ``Deep sparse rectifier
neural networks,''
  in \emph{Proc. Int. Conf. Artificial Intelligence and Statistics}, vol.~15,
  Fort Lauderdale, FL, 2011, pp. 315--323.

\bibitem{srivastava_dropout:_2014}
N.~Srivastava, G.~E. Hinton, A.~Krizhevsky, I.~Sutskever, and
R.~Salakhutdinov,
  ``{Dropout: A} simple way to prevent neural networks from overfitting.''
  \emph{J. Machine Learning Research}, vol.~15, no.~1, pp. 1929--1958, 2014.

\bibitem{glorot_understanding_2010}
X.~Glorot and Y.~Bengio, ``Understanding the difficulty of
training deep
  feedforward neural networks,'' in \emph{Proc. Int. Conf. Artificial
  Intelligence and Statistics}, vol.~9, Sardinia, Italy, 2010, pp. 249--256.

\bibitem{rumelhart_learning_1988}
D.~E. Rumelhart, G.~E. Hinton, and R.~J. Williams, ``Learning
representations
  by back-propagating errors,'' \emph{Nature}, vol. 323, pp. 533--536, 1986.

\bibitem{li2014efficient}
M.~Li, T.~Zhang, Y.~Chen, and A.~J. Smola, ``Efficient mini-batch
training for
  stochastic optimization,'' in \emph{Proc. ACM SIGKDD Int. Conf. Knowledge
  Discovery and Data Mining}, New York, NY, 2014, pp. 661--670.

\bibitem{kingma_adam:_2014}
D.~Kingma and J.~Ba, ``Adam: {A} method for stochastic
optimization,''
  \emph{arXiv preprint arXiv:1412.6980}, 2014.

\bibitem{song_tvsum:_2015}
Y.~Song, J.~Vallmitjana, A.~Stent, and A.~Jaimes, ``{TVSum:
Summarizing} web
  videos using titles,'' in \emph{Proc. IEEE Conf. Computer Vision and Pattern
  Recognition}, Boston, MA, 2015, pp. 5179--5187.

\bibitem{krizhevsky_imagenet_2012}
A.~Krizhevsky, I.~Sutskever, and G.~E. Hinton, ``Imagenet
classification with
  deep convolutional neural networks,'' in \emph{Proc. Int. Conf. Neural
  Information Processing Systems}, Lake Tahoe, NV, 2012, pp. 1097--1105.

\bibitem{ma_user_2002}
Y.-F. Ma, L.~Lu, H.-J. Zhang, and M.~Li, ``A user attention model
for video
  summarization,'' in \emph{Proc. ACM Int. Conf. Multimedia}, Juan les Pins,
  France, 2002, pp. 533--542.

\end{thebibliography}

\end{document}